\documentclass{ifacconf}

\usepackage{graphicx}      
\usepackage{natbib}        
\usepackage{siunitx}       
\usepackage{multirow}      

\begin{document}
\begin{frontmatter}

\title{Evaluating the Explainable AI Method Grad-CAM for Breath Classification on Newborn Time Series Data\thanksref{footnoteinfo1}\thanksref{footnoteinfo2}} 

\thanks[footnoteinfo1]{The project on which this report is based was funded by the Federal Ministry of Education and Research under the funding code 031L0303A.}
\thanks[footnoteinfo2]{© 2024 The authors. This work has been accepted to IFAC for publication under a Creative Commons Licence CC-BY-NC-ND.}

\author[First,Third]{Camelia Oprea}
\author[First,Third]{Mike Grüne}
\author[First]{Mateusz Buglowski}
\author[Second]{Lena Olivier}
\author[Second]{Thorsten Orlikowsky}
\author[First]{Stefan Kowalewski}
\author[Second,Fourth]{Mark Schoberer}
\author[First,Fourth]{André Stollenwerk}

\address[First]{Chair for Embedded Software, RWTH Aachen University, Aachen,
Germany (e-mail: oprea@embedded.rwth-aachen.de)}
\address[Second]{Neonatology Section of the Department of Paediatric and Adolescent Medicine, RWTH Aachen University Hospital, Aachen, Germany}
\address[Third]{Co-First Authors}
\address[Fourth]{Equally contributing}

\begin{abstract}                
With the digitalization of health care systems, artificial intelligence becomes more present in medicine. Especially machine learning shows great potential for complex tasks such as time series classification, usually at the cost of transparency and comprehensibility. This leads to a lack of trust by humans and thus hinders its active usage. Explainable artificial intelligence tries to close this gap by providing insight into the decision-making process, the actual usefulness of its different methods is however unclear. This paper proposes a user study based evaluation of the explanation method Grad-CAM with application to a neural network for the classification of breaths in time series neonatal ventilation data. We present the perceived usefulness of the explainability method by different stakeholders, exposing the difficulty to achieve actual transparency and the wish for more in-depth explanations by many of the participants.
\end{abstract}

\begin{keyword}
medical application, xAI, CNN, Grad-CAM, user study, time series, breath classification
\end{keyword}

\end{frontmatter}

\section{Introduction}
\label{sect:introduction}
Mechanical ventilation is a vital therapy for neonates with respiratory failure. Modern ventilators are able to not just deliver breaths with predetermined settings but can also detect the patient's breathing efforts and effectively support these. Time series data of flow and pressure thus exhibit specific patterns for different types of breaths. These allow the categorization of breaths, for example as spontaneous or mechanical.  The classification of breaths can be of interest for clinical and medical engineering applications, for instance during the weaning process of patients (\cite{sangsari2022weaning}). However, the classification of breaths by hand is a time-intensive task, requiring domain knowledge. Hence, an automated procedure is desirable. \cite{chong_computational_2021} introduced a rule-based approach for breath detection and division of breaths into phases for neonatal patients. The different breath types were however not the focus.

The neonatal breath data used in this work is recorded as multivariate times series. The utilization of artificial intelligence (AI) for time series classification constitutes a promising approach, considering its achievements for this task (\cite{ismail2019deep,ruiz2021great}).
In fields such as medicine, especially in the intensive care unit, the result of an AI-based system may have significant consequences. Thus, the comprehensibility of the underlying methodology plays an important part in the assessment of accountability and the user's confidence in the system. Achieving greater comprehensibility is the motivation behind the field of explainable artificial intelligence (xAI). Widespread xAI methods (\cite{ribeiro_why_2016, lundberg_unified_2017, selvaraju_grad-cam_2017}) have already been applied in medicine (\cite{fauvel_xcm_2021, raab2023xai4eeg}), their actual use in enhancing the explainability of the underlying method is however still questioned (\cite{Duran_Jongsma_2021}). To better asses the perceived explainability of xAI, the different stakeholders, that interact with the system, need to be regarded (\cite{arrieta_xai_2020}). These may have different demands and requirements with respect to the delivered explanations, from developers, aiming to improve the system, to end-users such as domain experts, aiming to successfully apply the system. 

This work presents the results of a user study based evaluation of a Grad-CAM implementation. As working example, the convolutional neural network (CNN) by \cite{fauvel_xcm_2021} is applied to breath classification in neonatal respiratory data. To evaluate the usefulness of the resulting visual explanations, two stakeholders are regarded: developers and domain experts.

\section{Data Description}
\label{sect:datadescr}

\begin{figure}
\begin{center}
\includegraphics[width=8.4cm]{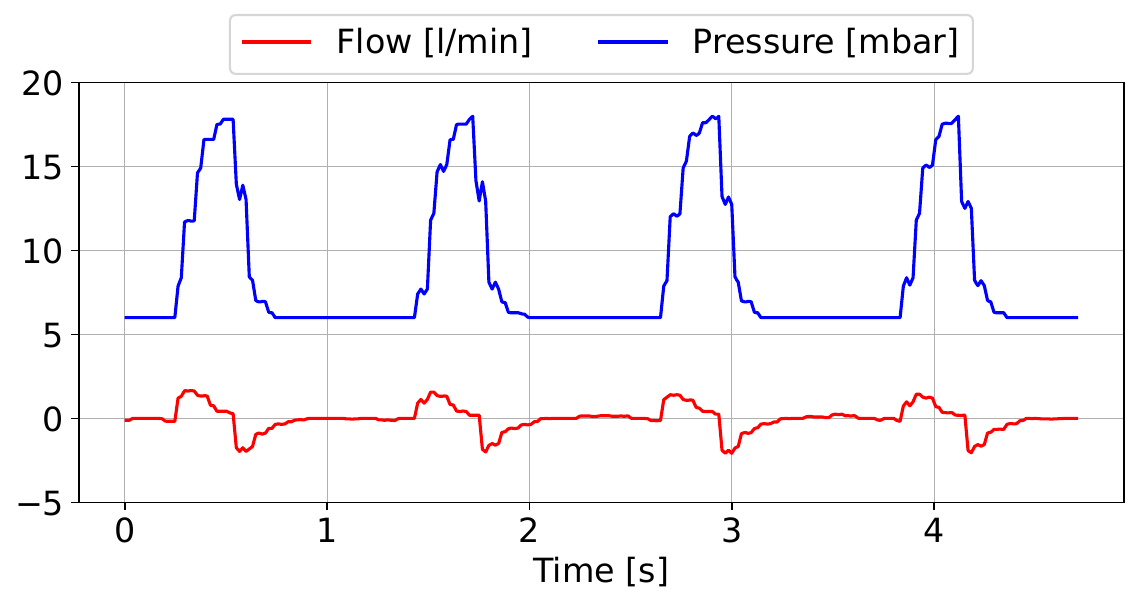}
\caption{Plotting of exemplary flow and pressure data.} 
\label{fig:fourBreaths}
\end{center}
\end{figure}

The used data was collected by the RWTH Aachen University Hospital and annotated by a senior neonatologist (M. Schoberer). The records comprise ventilation data and vital parameters from the neonatal intensive care unit. From these, the flow and pressure measurements during invasive mechanical ventilation are used for the proposed method of classifying single breaths. In total, $18$ patients are selected for the study. Both considered parameters are provided at a frequency of \SI{125}{\hertz} in each patient and the illustrative format of the data streams is depicted in Fig.~\ref{fig:fourBreaths}.

\begin{figure}
\begin{center}
\includegraphics[width=8.4cm]{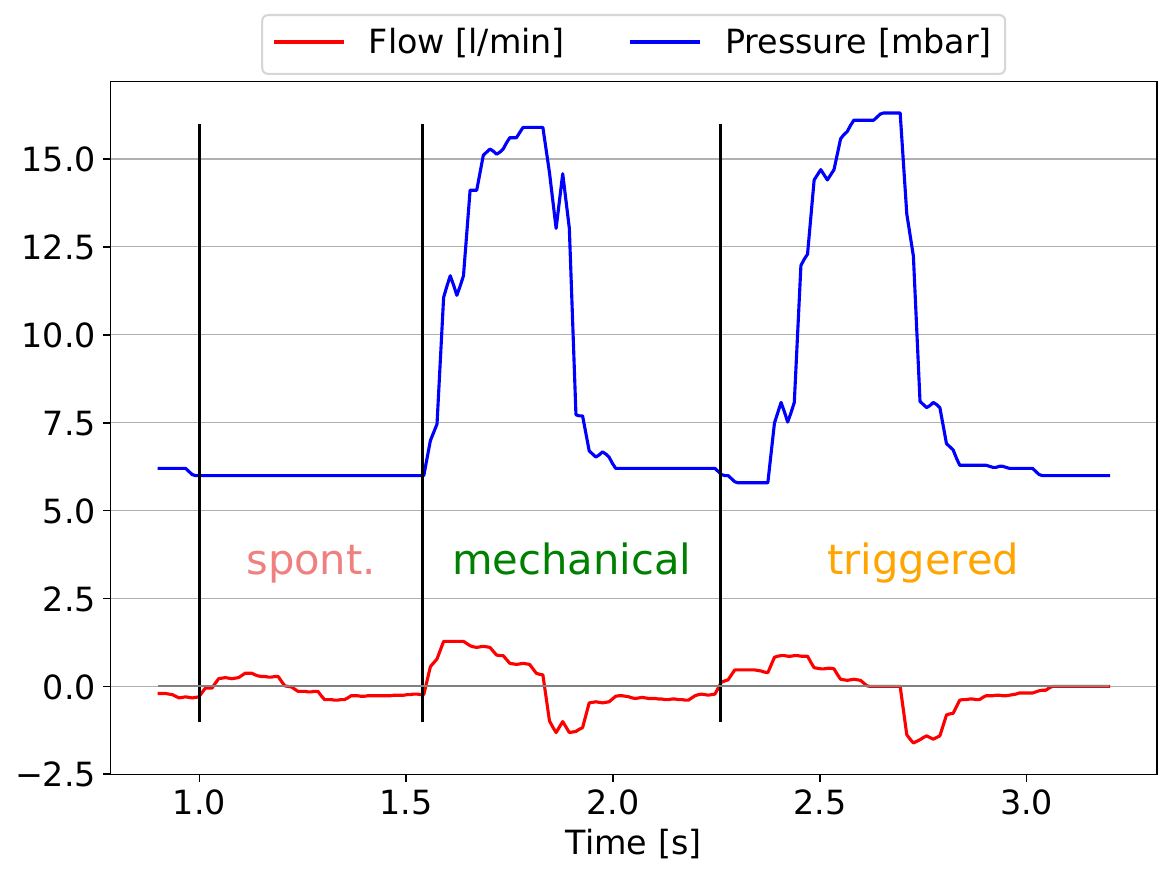}
\caption{Three of the identified breath types: spontaneous (spont.), mechanical and triggered. The unclassifiable category, which does not exhibit a specific representative form, is excluded here.} 
\label{fig:breathTypes}
\end{center}
\end{figure}

The utilized ventilator, while displaying some information on breath types, does not store the information about the classification of individual breaths. Hence, a zero crossing detection is used on the flow measurement to divide the time series data into time segments, representing potential breaths. 
This workflow is based on the fact, that a positive flow reflects air flowing into the lungs, whilst a negative flow represents air flowing out of the lungs. Thus, flow crossings from a negative to a positive value depict a potential start of an inspiration and, in the used definition of this work, also the end of the previous breath. Artefacts, such as flow oscillating around zero or single inspirations without accompanying expiration, do not represent an actual breath and should be recognized and distinguished by the proposed model as such. This is realised by including the artefacts in the annotation phase to be consequentially learned by the neural network.
The selected time segments are annotated and classified by the assisting neonatologist. For each of the 18 patients, five minutes per patient were extracted for annotation, resulting in a total of $6304$ potential breaths to be used for training and evaluation. Each potential breath, consisting of flow and pressure data, can either be classified into one of four breath types or can be annotated as an artefact and therefore not be categorized as a breath. In our definition, the following breath types exist: 
\begin{itemize}
    \item \textbf{Spontaneous}: the breath is fully carried out by the patient without the support of the ventilator.
    \item \textbf{Mechanical}: the ventilator initiates and sustains the breath.
    \item \textbf{Triggered}: the patient's inspiratory effort, creating positive airflow, triggers the ventilator.
    \item \textbf{Unclassifiable}: time segments, which are identified as breaths by the senior neonatologist, but exhibit characteristics, that prevent a clear assignment to the other classes.
\end{itemize}
Representative graphs for the breath types spontaneous, mechanical and triggered are depicted in Fig.~\ref{fig:breathTypes}. Unclassifiable breaths can by definition not be represented by a specific form.

The resulting annotated data, comprising 90 minutes and 18 patients, is subsequently split into distinct sets for the model training, validation and the final testing. From this dataset 15 minutes, each from a different patient, are randomly chosen as validation data. Furthermore, five minutes, each from a different patient, are randomly chosen as test dataset for the final evaluation, distinct of the training and validation data. The distribution of time segments per dataset and their annotations are described in Table~\ref{tb:annotatedData}.

\begin{table}[hb]
\addtolength{\tabcolsep}{-0.4em}
\begin{center}
\caption{Distribution of the annotated data for training, validation and the final testing.}\label{tb:annotatedData}
\begin{tabular}{c|cc|c}
    \textbf{Type} & Training & Validation & Test\\
    \hline
    No breath / Artefact & 1028 & 314 & 92\\
    Spontaneous breath & 807 & 199 & 36\\
    Mechanical breath & 1362 & 298 & 92\\
    Triggered breath & 1450 & 335 & 143\\
    Unclassifiable breath & 127 & 20 & 1\\
    \hline
    Total & 4774 & 1166 & 364
\end{tabular}
\end{center}
\end{table}

\section{Classification Model and Results}
\label{sect:methods}

\begin{figure*}[h]
\begin{center}
\includegraphics[width=18cm]{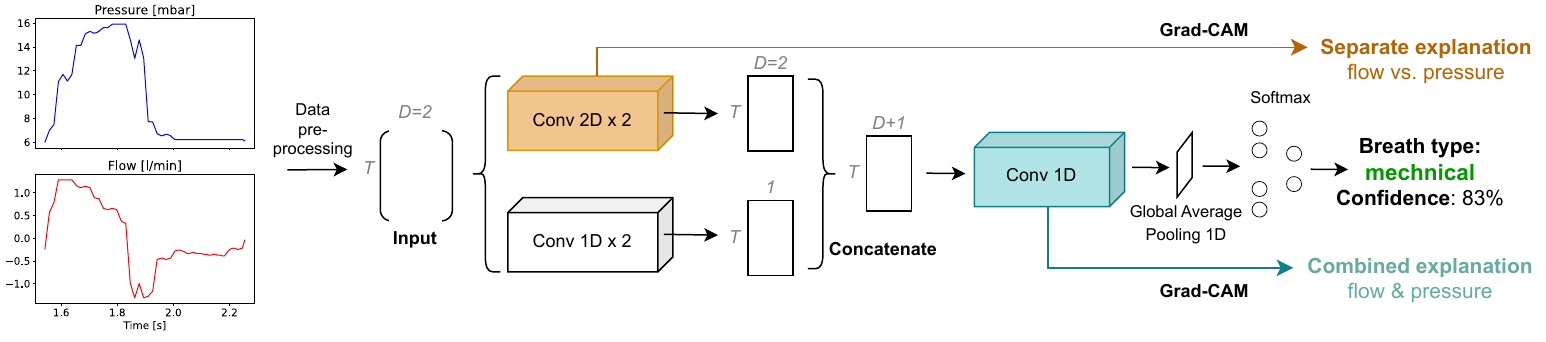}
\caption{Data pipeline from flow and pressure input to the resulting classification with a two-fold explanation. The CNN architecture is adopted from \cite{fauvel_xcm_2021}. Abbreviations: $D$ - number of observed variables, $T$ - time series length. Grad-CAM is applied to the 2D convolution (orange) to deliver variable-wise explanation and to the last 1D convolution (blue) to deliver a combined explanation.}
\label{fig:xcmArchitecture}
\end{center}
\end{figure*}

This chapter describes the trained network and gives an overview of its results, while chapter \ref{sect:evaluation} details the explainability aspect, which is the main focus of this paper.
The foundation for the proposed  breath classification method is a state-of-the-art explainable convolutional neural network for multivariate time series classification (XCM) proposed by \cite{fauvel_xcm_2021}. The network uses three convolutional layers, which bring a two-fold advantage. On the one hand, it enables the network to process the input data both separately and as the combination of all observed variables. This results in the extraction of more discriminative features. On the other hand, the approach allows a model-specific explanation, which can be applied to each of the underlying convolutional layers separately.

This facilitates the explanation of different stages in the network. The structure of our implementation of XCM for the use case of breath classification and the extraction of explanations is depicted in Fig.~\ref{fig:xcmArchitecture}.

The usage of convolutional layers requires a low amount of weights and enables a time-independent detection of features in time series data, which is argued to enhance the generalization ability in comparison to other state-of-the-art approaches~(\cite{fauvel_xcm_2021}). This property is valuable in the field of breath classification, as breaths show a high variability. The fixed input length of CNNs is however a limitation, considering the naturally varying breath duration and should be appropriately handled.

We use a time series length $T$ corresponding to \SI{5}{\second} of data, based on the maximal reference breath duration stated by \cite{baker_artificial_2020}. Most of the considered breaths are shorter than this and are thus bilaterally padded with zeros, to attain a data series of \SI{5}{\second}. Longer breaths are resampled in order to adhere to this maximum length. The used kernel size, defining the duration of data considered at once for feature detection, is determined by the best training result and is set to span \SI{344}{\milli\second}. \cite{fauvel_xcm_2021} use an additional padding and a sliding window stride of one. This avoids upsampling of the resulting feature maps, leading to a more accurate explanation generation, and attains features covering every part of the data. 

The training of the network is also adopted from the XCM implementation with a 5-fold cross-validation, a batch size of $32$ and $100$ epochs as hyperparameters. The model is evaluated on the dedicated test data, as mentioned in chapter \ref{sect:datadescr}. The results for the five randomly chosen patients with $61$, $52$, $39$, $55$ and $65$ annotated breaths respectively, are depicted in Table~\ref{tb:evaluationAccuracy}. From this we can observe the accuracy varying significantly between the chosen minutes of data. Furthermore, the model was prone to misclassifying artefacts as spontaneous breaths, which is reflected in the varying accuracy in Table~\ref{tb:evaluationAccuracy} when excluding annotated artefacts.

\begin{table}[h]
\addtolength{\tabcolsep}{-0.4em}
\begin{center}
\caption{Accuracy of the proposed ML model on test data. In the upper line, all detected breaths by the model are considered. In the bottom line, only actually annotated breaths are regarded.}\label{tb:evaluationAccuracy}
\begin{tabular}{c|ccccc}
    \textbf{\multirow{2}{*}{\shortstack{Considered\\ Breaths}}} & \multirow{2}{*}{Patient 1} & \multirow{2}{*}{Patient 2} & \multirow{2}{*}{Patient 3} & \multirow{2}{*}{Patient 4} & \multirow{2}{*}{Patient 5}\\
     & & & & & \\
    \hline
    \multirow{2}{*}{\shortstack{Including\\ Artefacts}} & \multirow{2}{*}{\SI{80.28}{\percent}} & \multirow{2}{*}{\SI{69.57}{\percent}} & \multirow{2}{*}{\SI{58.70}{\percent}} & \multirow{2}{*}{\SI{34.00}{\percent}} & \multirow{2}{*}{\SI{61.25}{\percent}}\\
     & & & & & \\
    \multirow{2}{*}{\shortstack{Excluding\\ Artefacts}} & \multirow{2}{*}{\SI{90.16}{\percent}} & \multirow{2}{*}{\SI{84.62}{\percent}} & \multirow{2}{*}{\SI{51.28}{\percent}} & \multirow{2}{*}{\SI{47.27}{\percent}} & \multirow{2}{*}{\SI{64.62}{\percent}}\\
     & & & & & \\
\end{tabular}
\end{center}
\end{table}

The metrics sensitivity, specificity and accuracy, depicted in Table~\ref{tb:evaluationMetrics}, reveal that the proposed model does not exhibit the desired generalisability for an unrestricted usage on additional data. The results show a weakness in the detection of mechanical breaths, indicated by the low sensitivity. An excessive classification of spontaneous and triggered breaths is shown by the specificity. Discrimination of unclassifiable breaths appears to be deficient, but the result must be regarded critically because only one suiting breath is included in the test dataset.

\begin{table}[h]
\addtolength{\tabcolsep}{-0.4em}
\begin{center}
\caption{Sensitivity, Specificity and Accuracy across the five minutes of test data.}\label{tb:evaluationMetrics}
\begin{tabular}{c|cccc}
    \textbf{Metric} & Spontaneous & Mechanical & Triggered & Unclassifiable\\
    \hline
    Sensitivity & \SI{100.00}{\percent} & \SI{46.74}{\percent} & \SI{75.52}{\percent} & \SI{0.00}{\percent}\\
    Specificity & \SI{77.58}{\percent} & \SI{93.43}{\percent} & \SI{78.03}{\percent} & \SI{97.81}{\percent}\\
    Accuracy & \SI{79.78}{\percent} & \SI{81.69}{\percent} & \SI{77.05}{\percent} & \SI{97.54}{\percent}\\
\end{tabular}
\end{center}
\end{table}

While the results do not come up with our expectations, the model should be seen as the basis for the explainability method, which constitutes the focus of this paper. The performance of the model is to be taken into account when interpreting the explainability evaluation results, while the possible advancements of the model are not thematized further.

\section{Explainability Method and Evaluation}
\label{sect:evaluation}

In the context of XCM, Grad-CAM (\cite{selvaraju_grad-cam_2017}) can be applied to each of the convolutional layers to obtain a layer-specific explanation. Grad-CAM allows generating heatmaps for the current inputs based on the trained weights and the resulting feature maps of a convolutional layer. In the use case of breath classification, this enables to determine and visualize the importance of each considered flow and pressure data point for the current classification, in separate or as combination. For the proposed model, we decided to apply Grad-CAM on the two-dimensional and the concluding one-dimensional convolutional layer of XCM, as shown in Fig.~\ref{fig:xcmArchitecture}. The obtained heatmaps thus comprise information about the separate variables, flow and pressure, in the two-dimensional layer and the combination of the two input variables in the concluding layer. The latter convolutional layer encodes  higher-level features than the preceeding one-dimensional convolution, therefore we regard its explanation as the most appropriate  for practical usage. The obtained heatmap of the separately explained inputs is made optional to the user. It does not necessarily represent important information for the final classification due to the design of XCM.

\begin{figure}[h]
\begin{center}
\includegraphics[width=8.4cm]{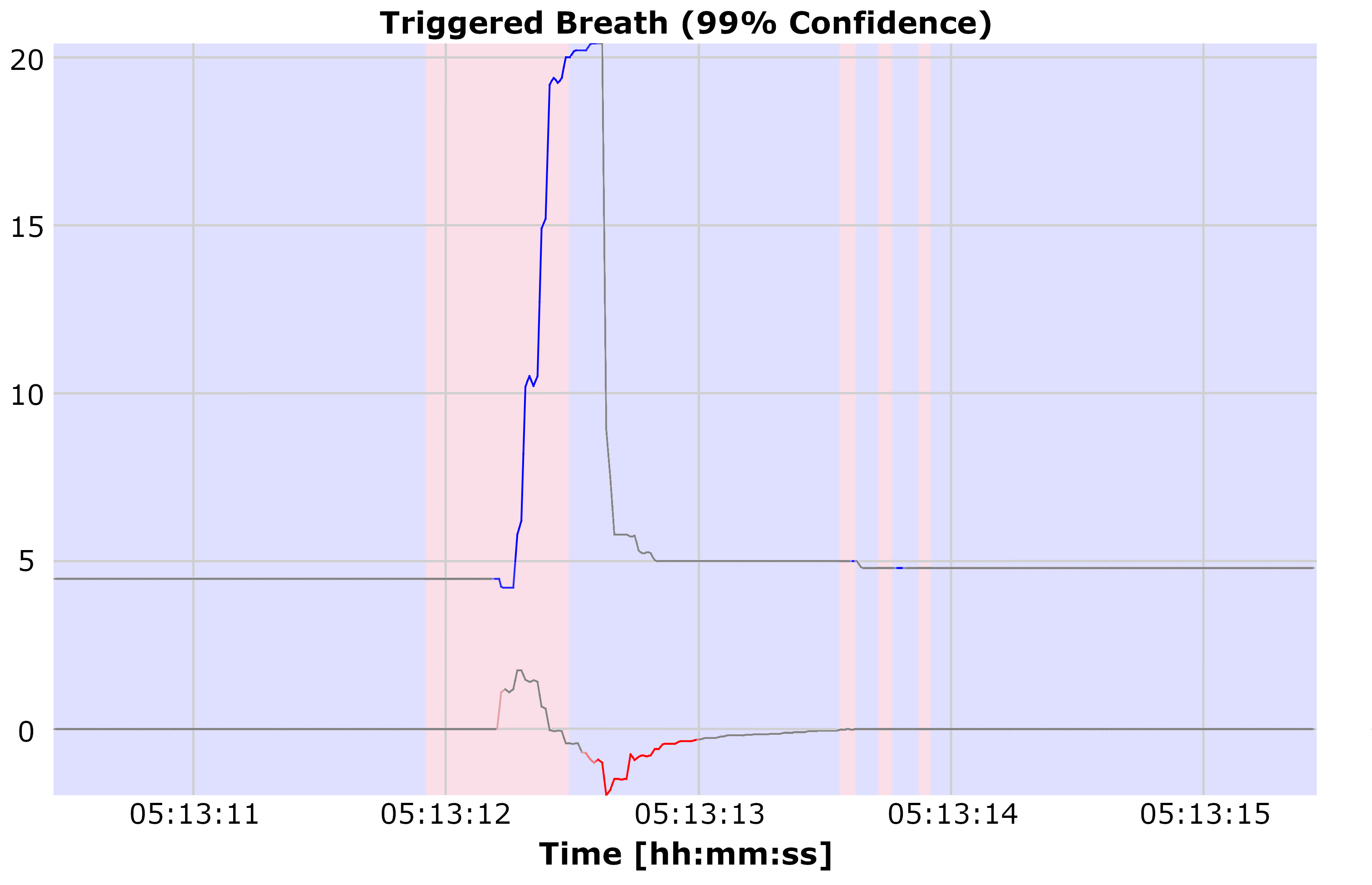}
\caption{Explanations of the classification as displayed in the used data viewer. The background color expresses the importance of the inputs' combination with a color gradient from low (blue) to high (red). The inputs' separate significance is depicted by the color intensity of the corresponding flow (red) and pressure (blue) curves. Both curves are visually extended with the first (05:13:12.2) and last (05:13:13.8) available value respectively for better perceptibility.} 
\label{fig:explanationsPlugin}
\end{center}
\end{figure}

For the evaluation, a self-developed data viewer is used, in which both emerging explanations are displayed, as depicted in Fig.~\ref{fig:explanationsPlugin}.  While the selected breath is processed by the network with a zero padding, the plotting of leading and trailing zeros leads to a confusing visualization. Therefore, in the data viewer the breath is visually padded with the first and last available value. The input's combined explanation of the final convolutional layer is presented as a heatmap in the background. The input's separate importance is visualized by the color intensity of the variable's corresponding graph. The classification of the network is textually displayed above the plot together with the related confidence of the network. The latter value is intended to further support the assessment of the classification of the current breath.

The usefulness of the proposed method is evaluated with the help of a self-developed questionnaire. The application-specific questions are based on relevant aspects of explainable artificial intelligence, as proposed by \cite{arrieta_xai_2020}. For this paper, the following facets are of special interest:
\begin{itemize}
    \item \textbf{Trustworthiness}: the model acts as intended when facing a problem.
    \item \textbf{Causality}: the explanation states correlations, which expectantly have a causal justification.
    \item \textbf{Transferability}: the explanations provide an assessment of the models generalisability limits.
    \item \textbf{Informativeness}: the explanations give enough information to grasp the models functionality.
    \item \textbf{Confidence}: the model and its explanations act robust and stable.
    \item \textbf{Fairness}: the model has not trained a bias.
\end{itemize}

The questionnaire is divided into three sections, which are conducted sequentially. First, the performance of the method is evaluated, followed by the proposed explanations and finally, the overall practicability is assessed. The three sections aim to cover the above mentioned xAI aspects. A precise assignment between a question and an aspect is however difficult. The perceived performance may cover trustworthiness, confidence but also fairness. The explanation evaluation may target the causality and informativeness, while the overall practicability might hint at the transferability of the chosen method. 
Each included question features a numerical answer ranging from $1$ (worst result) to $6$ (best result), which, by providing an even amount of possible answers, aims to prevent a tendency to the middle. This paper brings an evaluation on the explainability of a machine learning method with application in the field of neonatal ventilation. Therefore, the participants are chosen to cover both the medical and computer science field to include possible end users as well as developers of such algorithms. From the twelve participants, seven are medical professionals from the RWTH Aachen University Hospital. These represent the domain experts, as they are working or researching in the field of neonatal and pediatrics intensive care treatment or are experienced in mechanical ventilation. The remaining five participants, representing the developers, are scientists associated with the Chair for Embedded Software of RWTH Aachen University. They either work with algorithms on neonatal respiratory data or specialize in the field of explainable artificial intelligence. The results of the performed questionnaire will be considered separately for both groups of participants, to assess the difference in usefulness for the two very different fields of interest.

The participants were presented with the same continuous hour of data, randomly chosen from one of the 18 patients and distinct of the annotated data. Using the self-developed viewer, they could apply the proposed model in this timeframe of an hour without any restrictions. A tutorial about the usage is provided to all participants and the data viewer was already used by most of the participants beforehand. In order to establish a common ground, the participants were also presented with a simplified explanation of machine learning and the process of supervised learning for this specific use case. After the usage for an arbitrary amount of time, the participants were requested to fill out the questionnaire. 

\begin{figure}[h]
\begin{center}
\includegraphics[width=8.4cm]{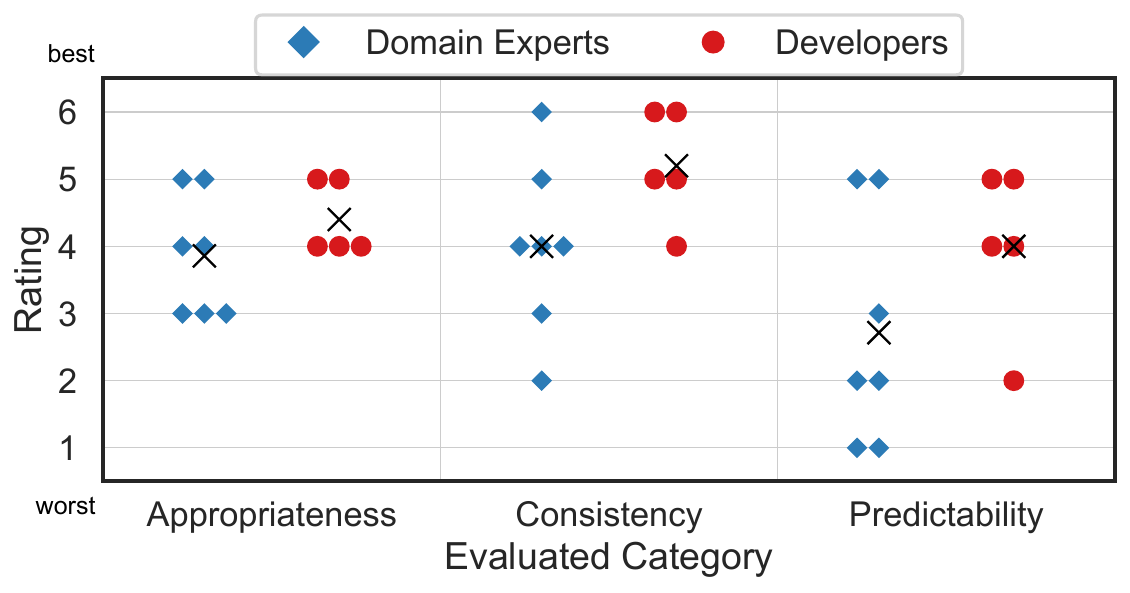}
\caption{Numeric results of the assessment of the perceived performance metrics separated by the participants' profession. The ratings go from 1 (worst) to 6 (best), with the mean rating per group being visualized as a cross.} 
\label{fig:performanceBoxplot}
\end{center}
\end{figure}

To assess the perceived performance of the algorithm, three performance aspects are evaluated: the appropriateness of the performed classifications, characterizing the agreement of the classifications with the users' assessment, the consistency of the classifications, assessing whether similar breaths are handled similarly by the model, and the predictability of the model, describing how foreseeable the classifications are. The obtained assessments are depicted in Fig.~\ref{fig:performanceBoxplot}. The results of this performance evaluation are inherently subjective and thereby enable an assessment with regard to different strategies and levels of knowledge in breath classification. These factors can have a great influence on the perception of the proposed explanations and the consequential assessment of the complete method. 

\begin{figure}
\begin{center}
\includegraphics[width=8.4cm]{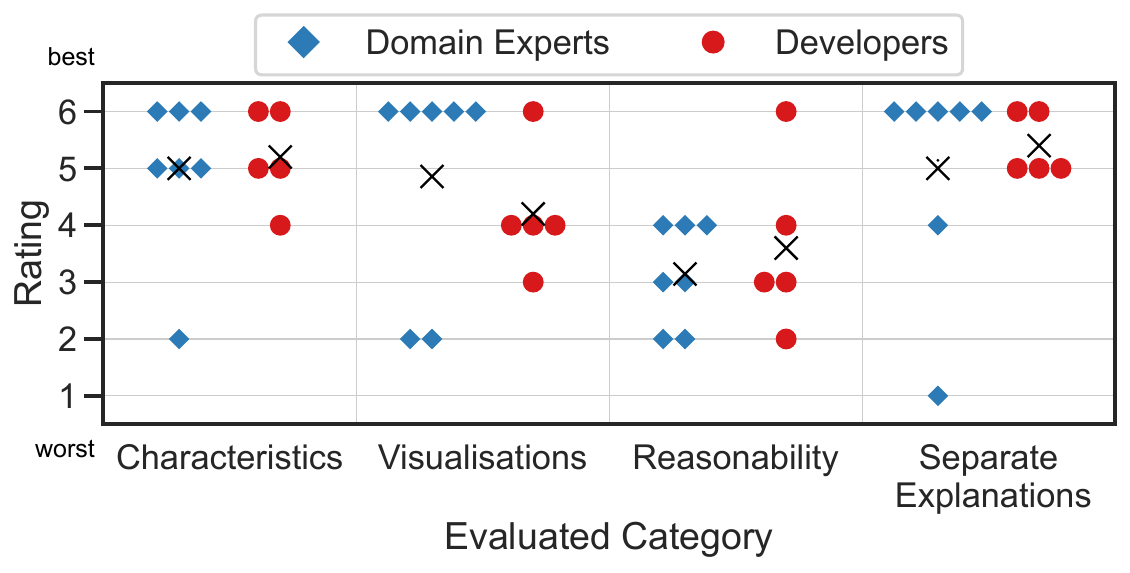}
\caption{Numeric results of the assessment of the perceived explainability metrics separated by the participants' profession. The ratings go from 1 (worst) to 6 (best), with the mean rating per group being visualized as a cross.}
\label{fig:explainabilityBoxplot}
\end{center}
\end{figure}

The implemented explanation methods are of particular interest to the conducted questionnaire, especially in the context of potential clinical usage by medical professionals. For this, four different questions of interest are defined.
\begin{enumerate}
    \item Does the explanation clearly point out, which characteristics of the breaths are relevant or irrelevant to the taken classification?
    \item Is the chosen visualization useful in understanding the resulting classification? 
    \item Does the given explanation display breath characteristics, which are considered reasonable?
    \item Are the implemented separate explanations for both flow and pressure assessed as useful by the user?
\end{enumerate}
The participants were asked to assess their impression for each question based on the hour of data presented for evaluation and their interaction with it. It was permitted to interrupt the questionnaire at any time and work on the data again if uncertainties arose. The results of this evaluation are depicted in Fig.~\ref{fig:explainabilityBoxplot}.

\begin{figure}
\begin{center}
\includegraphics[width=8.4cm]{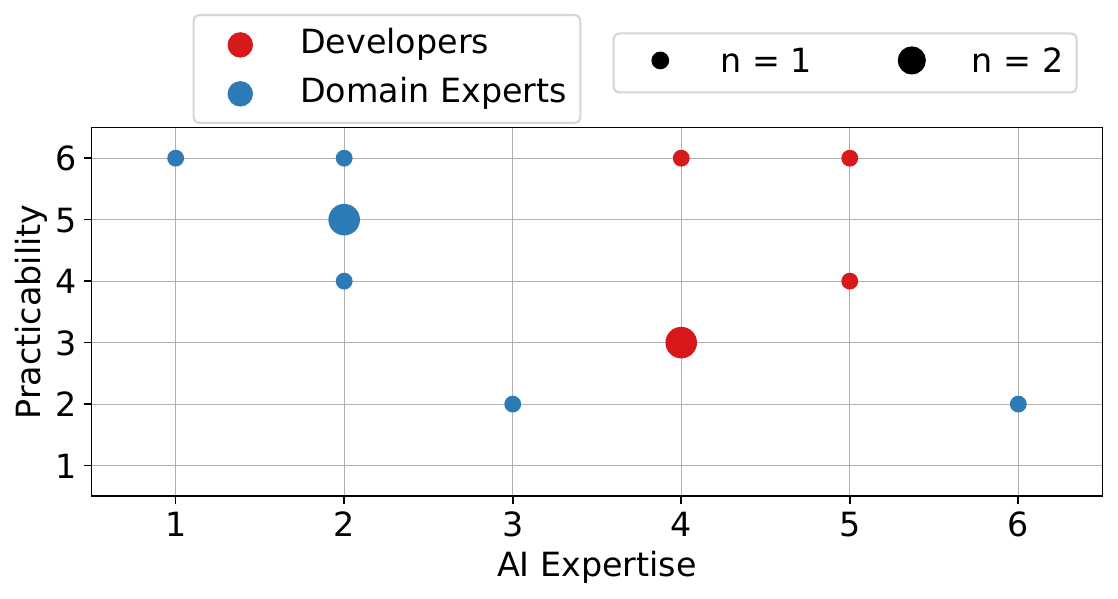}    
\caption{Scatterplot of the practicability assessment in correlation with the self-assessed AI expertise separated by the participants profession.} 
\label{fig:practicabilityScatterplot}
\end{center}
\end{figure}

Finally, the participants assessed the overall practicability of the proposed model in its current design. The results in correlation with the self-assessed AI expertise of the participants are presented in Fig.~\ref{fig:practicabilityScatterplot}. In the group of medical professionals, the assessment seems to correlate negatively with the self-assessed AI expertise.

In addition to the numerical evaluation of the given questions of interest, the participants could elaborate their assessment and provide suggestions for further improvements by using a provided text field or verbally during the in-person evaluation. Through the resulting remarks together with the numerical evaluation, some practical insights on the model are gained, as discussed in the next chapter.

\section{Discussion}
\label{sect:discussion}

Regarding the performance of the model, the assessment is worse in the group of medical professionals in every considered category and especially the predictability of the classifications is assessed poorly, as seen in Fig.~\ref{fig:performanceBoxplot}. One possible interpretation is, that people with less experience in the given task, might be more prone to trusting the explanations, accepting them as valid and even trying to detect and learn a reasoning from it. On the other hand, the domain experts might use their own knowledge to challenge the delivered explanation and classification.\\
A major point of criticism in the provided feedback, is the misclassification of spontaneous breaths. A spontaneous breath is performed by the patient, without the ventilator's support and thus is lacking a significant rise in pressure, see Fig.~\ref{fig:breathTypes}. This is a feature which was easily understood by both groups as a straightforward logic to distinguish spontaneous from triggered and mechanical breaths. Therefore, a breath classified as spontaneous, despite distinct indications of ventilator intervention, had an intensified misleading effect, due to the clarity of the human reasoning. Furthermore, the classification of artefacts as spontaneous breaths is rated as a problem of the proposed model. From the developers perspective it is troubling, that the algorithm is not capable to distinguish between random noise and a structured form. From the medical perspective, an excessive detection of spontaneous breaths could lead to a false interpretation of the patient's activity. Lastly, the explanations were criticized for highlighting meaningful segments but yielding the wrong classification. This was especially the case for mechanical breaths wrongly classified as triggered.
These observations are coherent with the results of the performance evaluation in chapter~\ref{sect:methods}. In this case of repeated misclassification, there was no clear evidence whether the explanations delivered any clarity.

Concerning the explanations, the assessments of developers and domain experts do not vary substantially. The visualisations are generally considered as useful, the rationale of the explanations and the relation between the two provided visualisations are however criticized. The latter is reasoned to yield uncertainties in the interpretation of the results. The influence of the separate explanations on the final classification was not clear and thus abstruse to most users. The most prominent criticism of the proposed explanations and the underlying model is the seemingly arbitrariness of important regions in some breaths. In many breaths, the participants assessed the identified important regions as reasonable. However, in some cases regions, such as the required data padding, were identified as important. These might have a rational explanation in the used model, but are not coherent with the reasoning process of the participants. Lastly, the importance of the provided separate visualisations is emphasized, while the meaningfulness in this model is doubted and the missing relation to the joint classification is criticized.

Generally, the method was assessed to not being useful for clinical usage in the current form, while most participants see a vast potential, especially in the context of research. One important note is the desire for a more explicit explanation by most of the medical professionals, indicating the insufficiency of the given explanations for the application of breath classification in neonatal intensive care.

\textit{Limitations.}
The choice for a CNN-based classification brings the constraint of a fixed length input. To overcome this limitation, we used zero padding. This, however, led to more error room for the classifier and the explanations, as they sometimes focused on the padding, which is not deemed as reasonable and confused the participants. Further, due to the rather low number of participants, which were further split into two groups, there was no easily discernible rating trend within a group.

\section{Conclusion}
\label{sect:conclusion}
This work underlines the many factors that influence a human-based evaluation of explainable AI. The knowledge of the task performed by the AI as well as knowledge on AI itself may affect the perceived usability of the explanation methods. Domain knowledge may affect the trustworthiness, as exemplified here by the medical experts. The self-assessed AI-expertise, however, had no clear overall correlation with the model's practicability assessment.
Further, misclassifications accompanied by unhelpful explanations reduce the informativeness and, hence, lower the confidence in the model. This is especially the case for classifications, which are straightforward for a human, the model and explanations however are wrong or misleading, and thus lower the trust. Lastly, the choice of visualizing the importance of the input data once separately and once together was assessed as a useful concept. However, it also raised more questions about the model's workings, affecting its perceived comprehensibility.
While such conclusions can be drawn from both the numerical assessment and the given remarks, these leave room for interpretation and should rather be used as groundwork for a more in-depth evaluation than definitive findings. Further user studies should cover data from multiple subjects. Additionally, a larger participant group is desired, to better cover the interplay of a person's understanding of both the AI method and the medical application.


\bibliography{ifacconf}             

\begin{thebibliography}{12}
\providecommand{\natexlab}[1]{#1}
\providecommand{\url}[1]{\texttt{#1}}
\providecommand{\urlprefix}{URL }
\expandafter\ifx\csname urlstyle\endcsname\relax
  \providecommand{\doi}[1]{doi:\discretionary{}{}{}#1}\else
  \providecommand{\doi}{doi:\discretionary{}{}{}\begingroup
  \urlstyle{rm}\Url}\fi

\bibitem[{Arrieta et~al.(2020)Arrieta, D{\'\i}az-Rodr{\'\i}guez, Del~Ser,
  Bennetot, Tabik, Barbado, Garc{\'\i}a, Gil-L{\'o}pez, Molina, Benjamins
  et~al.}]{arrieta_xai_2020}
Arrieta, A.B., D{\'\i}az-Rodr{\'\i}guez, N., Del~Ser, J., Bennetot, A., Tabik,
  S., Barbado, A., Garc{\'\i}a, S., Gil-L{\'o}pez, S., Molina, D., Benjamins,
  R., et~al. (2020).
\newblock Explainable artificial intelligence ({XAI}): Concepts, taxonomies,
  opportunities and challenges toward responsible ai.
\newblock \emph{Information fusion}, 58, 82--115.
\newblock \doi{10.1016/j.inffus.2019.12.012}.

\bibitem[{Baker(2020)}]{baker_artificial_2020}
Baker, D.J. (2020).
\newblock \emph{Artificial {Ventilation}: {A} {Basic} {Clinical} {Guide}}.
\newblock Springer International Publishing, Cham.
\newblock \doi{10.1007/978-3-030-55408-8}.

\bibitem[{Chong et~al.(2021)Chong, Morley, and
  Belteki}]{chong_computational_2021}
Chong, D., Morley, C.J., and Belteki, G. (2021).
\newblock Computational analysis of neonatal ventilator waveforms and loops.
\newblock \emph{Pediatric Research}, 89(6), 1432--1441.
\newblock \doi{10.1038/s41390-020-01301-9}.

\bibitem[{Dur{\'a}n and Jongsma(2021)}]{Duran_Jongsma_2021}
Dur{\'a}n, J.M. and Jongsma, K.R. (2021).
\newblock Who is afraid of black box algorithms? on the epistemological and
  ethical basis of trust in medical ai.
\newblock \emph{Journal of Medical Ethics}, medethics--2020--106820.
\newblock \doi{10.1136/medethics-2020-106820}.

\bibitem[{Fauvel et~al.(2021)Fauvel, Lin, Masson, Fromont, and
  Termier}]{fauvel_xcm_2021}
Fauvel, K., Lin, T., Masson, V., Fromont, {\'E}., and Termier, A. (2021).
\newblock {XCM}: {An} {Explainable} {Convolutional} {Neural} {Network} for
  {Multivariate} {Time} {Series} {Classification}.
\newblock \emph{Mathematics}, 9(23), 3137.
\newblock \doi{10.3390/math9233137}.

\bibitem[{Ismail~Fawaz et~al.(2019)Ismail~Fawaz, Forestier, Weber, Idoumghar,
  and Muller}]{ismail2019deep}
Ismail~Fawaz, H., Forestier, G., Weber, J., Idoumghar, L., and Muller, P.A.
  (2019).
\newblock Deep learning for time series classification: a review.
\newblock \emph{Data mining and knowledge discovery}, 33(4), 917--963.
\newblock \doi{10.1007/s10618-019-00619-1}.

\bibitem[{Lundberg and Lee(2017)}]{lundberg_unified_2017}
Lundberg, S. and Lee, S.I. (2017).
\newblock A {Unified} {Approach} to {Interpreting} {Model} {Predictions}.
\newblock \doi{10.48550/ARXIV.1705.07874}.

\bibitem[{Raab et~al.(2023)Raab, Theissler, and Spiliopoulou}]{raab2023xai4eeg}
Raab, D., Theissler, A., and Spiliopoulou, M. (2023).
\newblock {XAI4EEG}: spectral and spatio-temporal explanation of deep
  learning-based seizure detection in eeg time series.
\newblock \emph{Neural Computing and Applications}, 35(14), 10051--10068.
\newblock \doi{10.1007/s00521-022-07809-x}.

\bibitem[{Ribeiro et~al.(2016)Ribeiro, Singh, and Guestrin}]{ribeiro_why_2016}
Ribeiro, M.T., Singh, S., and Guestrin, C. (2016).
\newblock "{Why} {Should} {I} {Trust} {You}?": {Explaining} the {Predictions}
  of {Any} {Classifier}.
\newblock In \emph{Proceedings of the 22nd {ACM} {SIGKDD} {International}
  {Conference} on {Knowledge} {Discovery} and {Data} {Mining}}, 1135--1144.
  ACM, San Francisco California USA.
\newblock \doi{10.1145/2939672.2939778}.

\bibitem[{Ruiz et~al.(2021)Ruiz, Flynn, Large, Middlehurst, and
  Bagnall}]{ruiz2021great}
Ruiz, A.P., Flynn, M., Large, J., Middlehurst, M., and Bagnall, A. (2021).
\newblock The great multivariate time series classification bake off: a review
  and experimental evaluation of recent algorithmic advances.
\newblock \emph{Data Mining and Knowledge Discovery}, 35(2), 401--449.
\newblock \doi{10.1007/s10618-020-00727-3}.

\bibitem[{Sangsari et~al.(2022)Sangsari, Saeedi, Maddah, Mirnia, and
  Goldsmith}]{sangsari2022weaning}
Sangsari, R., Saeedi, M., Maddah, M., Mirnia, K., and Goldsmith, J.P. (2022).
\newblock Weaning and extubation from neonatal mechanical ventilation: an
  evidenced-based review.
\newblock \emph{BMC Pulmonary Medicine}, 22(1), 1--12.
\newblock \doi{10.1186/s12890-022-02223-4}.

\bibitem[{Selvaraju et~al.(2017)Selvaraju, Das, Vedantam, Cogswell, Parikh, and
  Batra}]{selvaraju_grad-cam_2017}
Selvaraju, R.R., Das, A., Vedantam, R., Cogswell, M., Parikh, D., and Batra, D.
  (2017).
\newblock Grad-{CAM}: {Why} did you say that?
\newblock \doi{10.48550/arXiv.1611.07450}.

\end{thebibliography}

\end{document}